\documentclass[sigconf]{acmart} 

\usepackage[normalem]{ulem} 

\newif\ifshowchanges
\showchangesfalse   

\ifshowchanges
  
  \newcommand{\deleted}[1]{\textcolor{red}{\sout{#1}}}
  
  \newcommand{\note}[1]{\textcolor{magenta}{[NOTE: #1]}}
\else
  
  \newcommand{\deleted}[1]{}
  
  \newcommand{\note}[1]{}
\fi

\usepackage[table]{xcolor} 
\usepackage{colortbl}      

\usepackage{lipsum}
\usepackage{multirow}

\newcommand{\nsd}[2]{\ensuremath{{#1}^{ \pm #2}}}

\newcommand{\qwk}[3][]{\ensuremath{{#2}^{\pm #3}_{#1}}}

\usepackage{soul}
\usepackage{lipsum}

\usepackage{makecell}   
\AtBeginDocument{%
  }

\setcopyright{none}
\renewcommand\footnotetextcopyrightpermission[1]{}

\begin{document}

\title{GradeLegal: Automated Grading for German Legal Cases}
\titlenote{This paper is an extended version of a short paper accepted at the International Conference on Artificial Intelligence and Law (ICAIL 2026).}


\author{Abdullah Al Zubaer}
\authornote{Corresponding author.}
\email{abdullahal.zubaer@uni-passau.de}
\orcid{0009-0001-0842-7434}
\affiliation{%
  \institution{University of Passau}
  \city{Passau}
  \country{Germany}
}

\author{Lorenz Wendlinger}
\email{lorenz.wendlinger@th-deg.de}
\orcid{0000-0001-9459-6244}
\affiliation{%
  \institution{Deggendorf Institute of Technology}
  \city{Deggendorf}
  \country{Germany}
}

\author{Simon Alexander Nonn}
\email{simon.nonn@uni-passau.de}
\affiliation{%
  \institution{University of Passau}
  \city{Passau}
  \country{Germany}
}

\author{Michael Granitzer}
\orcid{0000-0003-3566-5507}
\email{michael.granitzer@uni-passau.de}
\affiliation{%
  \institution{University of Passau}
  \city{Passau}
  \country{Germany}
}

\author{Jelena Mitrovi\'c}
\orcid{https://orcid.org/0000-0003-3220-8749}
\email{jelena.mitrovic@uni-passau.de}
\affiliation{%
  \institution{University of Passau}
  \city{Passau}
  \country{Germany}
}


\renewcommand{\shortauthors}{Abdullah Al Zubaer et al.}

\begin{abstract}
Grading German legal exam solutions faces growing volumes and a shortage of qualified graders, delaying feedback and creating a bottleneck.
At the same time, it is a high-stakes expert task, since state exam grades strongly influence career outcomes in Germany.
Despite this practical relevance, literature lacks systematic studies on effective methods for grading legal exams.
To address this gap, we investigate whether large language models (LLMs) can support the automated grading of German legal case solutions in criminal and public law, thereby enabling scalable feedback and student self-testing.
We present a systematic evaluation of 27 proprietary and open-source LLMs, benchmarking prompting strategies that incrementally add task-related information, such as a sample solution and a grading rubric.
Using quadratic weighted kappa (QWK), reasoning-oriented LLMs can approximate expert grading in public law when given a sample solution and a grading rubric (up to 0.91), compared to 0.60 in criminal law, suggesting a harder grading task in criminal law.
Beyond single-model grading, ensembling improves agreement by up to 0.15 over its best member and can offer an alternative to stronger closed-source single models.
In addition, our findings suggest that effective prompt design and model selection are necessary for reliable LLM-based grading of legal exams.
\end{abstract}

\ccsdesc[500]{Applied computing~Law}
\ccsdesc[500]{Computing methodologies~Natural language processing}
\ccsdesc[300]{Social and professional topics~Student assessment}

\keywords{Automated Essay Scoring, Large Language Model, German Legal Education, Legal Exam Assessment, Prompt Engineering }


\maketitle

\section{Introduction}

Examination grading is a core task in legal education that is central to obtaining the law qualification. For example, Germany's state exams include multiple written examinations graded by legal experts~\cite{strecker2026ki}. In these exams, answers are graded on the German 18-point ordinal scale (0--18)~\cite{FUberlinLaw_grading_scales}.
This process is time-consuming and may take weeks to months before grades are released. Key culprits are the complexity of the legal domain and, consequently, the student's answers, as well as limited expert capacity. 
In Germany, about 115,000 law students take written examinations each year, resulting in thousands of answers to grade~\cite{strecker2026ki}.
Given limited expert capacity, grading is often outsourced to external graders, which increases costs.
Grading is further complicated because the same conclusion can be reached through different legally coherent lines of reasoning. These paths must be evaluated for legal correctness, internal consistency, and argumentative soundness across the full solution. These characteristics complicate automation and distinguish legal exam grading from many short-form educational scoring settings.

Recent advances in large language models (LLMs) have demonstrated strong performance on legal reasoning tasks~\cite{yu-etal-2023-exploring, yuan-etal-2024-large, servantez-etal-2024-chain}.
This raises the question of whether LLMs can support automated grading for legal examinations, i.e., automated essay scoring (AES), and which prompting strategies enable them to approximate expert judgments.
If their grading performance aligns with expert judgments, LLM-based grading could enable scalable student self-assessment to estimate their performance on practice case solutions.
While AES has a long tradition in educational assessment~\cite{li-ng-2024-automated, Ramesh2022}, most prior work studies non-legal domains and relatively short responses~\cite{li-ng-2024-automated, Ramesh2022, ormerod-kehat-2025-long}.
Applicability of AES to complex, long-form legal examinations remains underexplored, particularly in the German legal domain, with only a few recent attempts~\cite{wendlinger2024suitabilitypretrainedfoundationalllms, strecker2026ki}.
These existing studies report mixed results across subdomains --- near-zero agreement in criminal law but improved grading in public law---yet the evidence is based on limited model coverage and evaluation setups. As a result, it remains unclear which combinations of model choice, prompt design, and feasibility of AES using LLMs in German legal examinations.
Moreover, beyond single-model grading, the potential of multi-model approaches such as ensembling to increase grading reliability in this setting has not been evaluated.

In this work, we study LLM-based grading for German criminal and public law examinations on an existing non-public dataset. We systematically evaluate 27 state-of-the-art LLMs reasoning and non-reasoning, open- and closed-source across multiple prompting strategies that incrementally add relevant task information (e.g., sample solutions and explicit rubrics aligned to the 18-point scale).
Beyond selecting a single {model~\cite{fan2025lexambenchmarkinglegalreasoning}, we test whether ensemble grading can outperform the best member model.}
Using quadratic weighted kappa (QWK), we measure LLM-human agreement with expert-graded exam answers to quantify when and how LLMs can approximate expert grading. Our contributions are the following: 

\begin{enumerate}
    \item To the best of our knowledge, we provide the first comprehensive, systematic evaluation of proprietary and open LLMs, covering reasoning and non-reasoning variants, and include 27 models evaluated on 87 spanning 71 criminal and 16 public law student case solutions.
    \item We benchmark prompting strategies that vary in task information (e.g., sample solution, rubric) on a non-public exam corpus to mitigate data contamination concerns~\cite{sainz-etal-2023-nlp}, and derive practical guidance for prompt design and model selection in legal exam grading.
    \item We show that recent reasoning LLMs, GPT-5, can approximate expert grading when given a sample solution and a grading rubric, achieving a QWK of up to 0.911 in public law, matching human agreement. Criminal law reaches a substantial agreement of up to 0.599, consistent with higher task complexity. Both exceeding previously reported results~\cite{wendlinger2024suitabilitypretrainedfoundationalllms, strecker2026ki}.
    \item We show that ensembles of open grader models outperform their best individual member under identical prompting---most strongly with full-context prompting---and narrow the gap to our strongest closed model.
\end{enumerate}

\section{Related Work}
We provide an overview of related work on LLMs for legal tasks, automated essay scoring, and LLM-based grading in legal education.
\subsection{LLMs in Legal NLP and Legal Reasoning Benchmarks}

LLM-driven legal benchmarks primarily assess legal reasoning capabilities rather than grade examination solutions.
LegalBench~\cite{10.5555/3666122.3668037} evaluates legal reasoning skills, and~\cite{10.1145/3709025.3712219} introduces a reasoning-centered legal retrieval augmented generation benchmark for evaluating LLM-based systems.
Other benchmarks are emerging that test legal knowledge and reasoning in examination-like or bar exam settings, such as GreekBarBench~\cite{chlapanis-etal-2025-greekbarbench}, LEXam~\cite{fan2025lexambenchmarkinglegalreasoning}, Brazilian Bar Exam~\cite{10.1145/3769126.3769227}, LawBench~\cite{fei-etal-2024-lawbench}, LaborBench~\cite{10.1145/3769126.3769256}.
Additionally, LEXam~\cite{fan2025lexambenchmarkinglegalreasoning} reports that ensembling model outputs (e.g., minimum score ensembles) can improve scoring performance in evaluation. We also adopt this ensemble insight in our study.
Beyond benchmarking, legal NLP research has addressed a broad range of tasks, including employment-contract clauses classification~\cite{10.1145/3769126.3769209}, legal argument mining~\cite{al2023performance}, legal document classification~\cite{hakimi-parizi-etal-2023-comparative}, legal question answering in the German civil law system~\cite{buttner-habernal-2024-answering}, and legal judgment prediction~\cite{wu-etal-2023-precedent}. While these benchmarks and related legal NLP tasks test legal knowledge and reasoning, they do not directly evaluate alignment with expert grading on exam-style legal case solutions.

\subsection{Automated Essay Scoring and LLM-based Grading}


Automated essay scoring (AES) has been widely explored in educational assessment~\cite{li-ng-2024-automated, Ramesh2022}. Traditional AES systems rely on classical machine learning or deep learning models with handcrafted features and require large labeled datasets for training or fine-tuning~\cite{wang-etal-2022-use, Uto2021, xie-etal-2022-automated}.
Many common AES benchmarks contain relatively short essays (e.g., ASAP~\cite{asap-aes} and ASAP 2.0~\cite{CROSSLEY2025100954,ormerod-kehat-2025-long}), unlike legal case solutions that normally span 10-30 handwritten pages~\cite{wendlinger2024suitabilitypretrainedfoundationalllms}. This gap in length and structure is substantial, since model architectures and evaluation protocols developed for short essays may not transfer directly to long-form legal cases.

AES research has explored fine-tuning pre-trained encoder models~\cite{ormerod-kehat-2025-long, 11062635}, such as BERT~\cite{devlin-etal-2019-bert} and Longformer~\cite{beltagy2020longformerlongdocumenttransformer}. Wang et al.~\cite{wang-etal-2022-use} reported improved performance for the ASAP and CRP datasets over earlier architectures (e.g., LSTM). Xie et al.~\cite{xie-etal-2022-automated} achieve state-of-the-art performance on ASAP using a BERT-based encoder with pairwise contrastive regression (0.817 QWK). 
Ormerod and Kehat~\cite{ormerod-kehat-2025-long} fine-tuned recent language models, including ModernBERT, Longformer, and Llama 3.2-8B, on the ASAP 2.0 dataset (with 17K/7.5K train/test instances). They aimed to address the limitation that long essays often exceed the input length of traditional models such as BERT. Results show that fine-tuned models can outperform a human baseline (0.75), with Longformer reaching 0.80. 
Overall, while fine-tuning encoder-based models can perform well on standard AES benchmarks, including longer responses, it typically requires substantial labeled data, compute, and technical expertise. Therefore, limiting out-of-the-box use in educational and professional settings, especially for course instructors who want to adopt these models as is. Moreover, much of this evidence comes from non-legal AES datasets rather than from legal-exam grading tasks.


Recent LLM-based approaches enable AES via prompting rather than model training and can achieve competitive performance in certain domains~\cite{huang-wilson-2025-evaluating}. Their extended context windows (in contrast to BERT or Longformer), together with in-context learning~\cite{NEURIPS2020_1457c0d6}, also make them natural candidates for long-form responses. At the same time, prompting-based scoring is sensitive to the information provided (e.g., rubrics, reference answers), thereby motivating controlled comparisons of prompting strategies~\cite{11272818, huang-wilson-2025-evaluating, golchin-etal-2025-grading}.

Seßler et al.~\cite{10.1145/3706468.3706527} examine whether LLMs can grade German school-level essays and report significant correlations between teacher ratings and proprietary reasoning-capable LLMs, while open-source non-reasoning models show weaker correlations. Their study provides evidence that LLM-based scoring can work in German language educational settings. 
Golchin et al.~\cite{golchin-etal-2025-grading} evaluate whether LLMs can replace peer grading in online courses. Using zero-shot chain-of-thought prompting, they compared three inputs: instructor answers only, instructor answers with an instructor rubric, and instructor answers with an LLM-generated rubric. 
Across three English-language courses using GPT-4 and GPT-3.5, the instructor-answer plus rubric setting, especially with GPT-4, aligned most closely with instructor grades and outperformed peer grading.
Liew and Tan~\cite{10.1145/3709026.3709030} reported substantial agreement on IELTS essay datasets (0.68 QWK) using GPT-3.5. 
Huang and Wilson~\cite{huang-wilson-2025-evaluating} evaluate closed-source LLMs on U.S. grade 3-4 essays, varying prompting and contextual information. They show that adding relevant context substantially improves performance (0.27 QWK on average), with their best setting reaching 0.72 (vs. 0.76 for a proprietary system; human agreement 0.91). Mansour et al.~\cite{mansour-etal-2024-large} evaluate GPT-3.5 and Llama-2 on the full ASAP dataset and find that generative LLM prompting lags far behind ( 0.31 and 0.30 QWK) fine-tuned encoder-based state-of-the-art~\cite{xie-etal-2022-automated}. However, their experiments use earlier model generations and do not include more recent reasoning-focused or current-generation LLMs.

Taken together, prior work in AES outside the legal domain shows that LLM-based scoring correlates with human ratings and that providing rubrics or context can improve performance.
Existing studies often focus on non-legal domains or short responses, evaluate only a few (often proprietary, mostly based on OpenAI) models, or lack controlled comparisons of prompting strategies.
Whether these findings generalize to long-form legal exam grading remains uncertain.

\subsection{Automated Assessment in Legal Education}
Compared to general AES, research on automated grading in legal education remains limited. Only recently has work begun to investigate LLMs in legal exam contexts~\cite{wendlinger2024suitabilitypretrainedfoundationalllms,strecker2026ki,Cope2025GradingMachines}. In this work, we use the public and criminal law exam dataset introduced by Strecker et al.~\cite{strecker2026ki} and Wendlinger et al.~\cite{wendlinger2024suitabilitypretrainedfoundationalllms} (see Section~\ref{dataset_overview}).

Cope et al.~\cite{Cope2025GradingMachines} study whether off-the-shelf LLM (GPT-5) can approximate professors' grades on U.S. law school exams. Using 205 student submissions from four final exams, they compare LLM scores to instructor grades.
They find that a detailed grading rubric increases alignment (correlation up to 0.93), while grading without explicit guidance reaches up to 0.80.
Their results support the value of structured rubrics, but they are limited to English-language U.S. exams and a single model family.

Wendlinger et al.~\cite{wendlinger2024suitabilitypretrainedfoundationalllms} evaluate pre-trained LLMs for grading in German legal education, focusing on criminal law with Llama-3 and Mixtral. Using few-shot prompting, they report near-zero (best$\approx$0.058) or negative correlations and conclude that criminal law exam grading is too complex for LLMs. They test several simplifications (additive scoring, partial credit, and chain-of-thought prompting) but observe no gains. However, they do not provide a sample solution or a grading rubric, which may limit the achievable level of agreement. Earlier work~\cite{11272818} suggests that such task-specific information can improve LLM-based scoring, motivating our addition of an instructor-written sample solution and a simplified rubric prepared by a legal expert.

Strecker et al.~\cite{strecker2026ki} study LLM-assisted grading for German public law case exams on the 18-point ordinal scale in a two-phase design. In the second phase, they select and analyze 16 digitally written student answers and examine, among didactic issues regarding legal grading, whether (1) an explicit rubric reduces variance among expert human graders and (2) LLMs can support or enhance grading under these conditions. They also compare human to LLM-generated feedback with regard to formative and summative aspects.
For LLM-based grading, they compare two prompting pipelines: one that grades more holistically, and the other that structures grading more fine-grained along the solution.
They evaluate GPT-4o and Gemini 2.5 Pro for both these solutions with and without rubric guidance, using an instructor-written sample solution, resulting in four different feedback outcomes.
They primarily assess performance using mean absolute point deviation from the instructor's grade and report that Gemini-2.5-pro has the minimum deviation in the fine-grained pipeline.
To better reflect the ordinal nature of the 18-point scale, we additionally calculate QWK and similarly find that Gemini 2.5 Pro performs best on this dataset (0.764) when given both the rubric and the sample solution.

Prior work in legal-exam grading~\cite{Cope2025GradingMachines,wendlinger2024suitabilitypretrainedfoundationalllms,strecker2026ki} does not yet establish how grading rubric/sample solution effects vary across model families or legal domains in German long-form exams. We therefore evaluate closed- and open-source models in a controlled setting covering both criminal and public law.

\section{Methodology}
This section outlines the methodological design for studying LLM-based grading of German legal case solutions, covering the data, model selection, prompting strategies, and evaluation protocol.
LLM-based grading is framed as a zero-shot task without graded exemplars: given the case facts and question (Table~\ref{tab:case_facts_tasks}) and a student's written solution, the model assigns a grade on the German 0--18 scale.
To quantify how additional context affects performance, the amount of grading information provided to the model is varied.
Accordingly, prompting strategies are benchmarked that \textit{incrementally add task-relevant information}, in particular, a grading rubric and/or one instructor-written sample solution for the target case.
Four prompt settings with increasing context availability are compared: (i) \textit{Task-Agnostic}, (ii) \textit{Instr.+Rubric}, (iii) \textit{Instr.+Solution}, and (iv) \textit{Instr.+Rubric+Solution}.
All settings are evaluated across multiple LLMs, and agreement with the reference grades is reported.

\subsection{Dataset Overview}
\label{dataset_overview}

We choose two non-public datasets of German law case solutions written in appraisal style and graded on the German legal grading scale (0--18), with a passing threshold of 4: a take-home \textit{criminal law} dataset~\cite{wendlinger2024suitabilitypretrainedfoundationalllms} and a timed mock exam \textit{public law} dataset~\cite{strecker2026ki}. Because the datasets are non-public, it is less plausible that the evaluated exam answers appear in LLM training data~\cite{sainz-etal-2023-nlp}.

The first dataset covers \textit{criminal law}~\cite{wendlinger2024suitabilitypretrainedfoundationalllms}. German law students in intermediate and advanced courses received the same case file as a take-home exam and wrote their solutions over several months.
Given the extended time window and unrestricted research materials, the selected topic is more complex and may be more open-ended, spanning multiple issues. Student solutions can therefore involve longer reasoning chains and be longer in text, which likely increases task complexity relative to timed exam settings where only the law itself is available.
Table~\ref{tab:case_facts_tasks} provides an English translation of the case facts and tasks.
The dataset contains 71 submissions for this case, all graded by expert research assistants using a sample solution provided by the course instructor. For this study, we enrich the dataset by adding an instructor's sample solution, which was not available in the original experiment~\cite{wendlinger2024suitabilitypretrainedfoundationalllms}. The sample solution specifies the full-credit model answer. We also created a simplified grading rubric that specifies how scores are allocated. The rubric provides tabular guidance on which elements in a student's answer earn which points; points are summed to determine the final grade. The rubric uses 100 points, which are then converted to the German grading scale 0-18 via a conversion table. A legal expert developed the conversion table and the rubric based on the instructor's written guidance; unlike rubrics authored by the exam's course instructor (as in our second dataset), it may, however, not fully reflect the instructor's intended weighting.

The second dataset covers \textit{public law}~\cite{strecker2026ki} and is derived from a mock exam.
The dataset comprises 16 student submissions for a single public law case and spans the full German grading scale (see Table~\ref{tab:case_facts_tasks}). 
Students wrote under state exam conditions (5 hours) with limited research materials (law books only).
This setting restricts the set of admissible issues and typically yields simpler solutions than our take-home criminal law dataset.
The submissions were graded by the course instructor who designed the exam, using his solution and his grading rubric. We treat the instructor's grades as the reference for our evaluation and report all assessments against this.
They show that even for human graders, rubrics are essential; based on additional human evaluation, the final inter-rater reliability is 0.911.

Table~\ref{tab:dataset_stats} summarizes both datasets. Criminal law answers are about four times longer in tokens (and words) than public law answers, with higher token (word) variance, indicating more extensive written reasoning in criminal law.
Average grades are higher in criminal law than in public law, with similar variance.
The criminal law exam solution was curated to span the full grading spectrum, as reflected in its sparse distribution of grades. In contrast, public law grades are concentrated toward the lower end of the scale.
As noted in~\cite{FUberlinLaw_grading_scales}, grades of 16-18 points are unlikely for legal case solutions, which is also reflected in these datasets.

\begin{table}[t]
\centering
\caption{Summarized case facts and student tasks for criminal and public law; translated to English from German.}
\begin{tabular}{@{} p{0.65\linewidth} p{0.3\linewidth}@{}}
\hline
\textbf{Case fact} & \textbf{Student task} \\
\hline

\textbf{Criminal Law:} A altered an inspection sticker, obstructed an overtaking driver who crashed while avoiding collision, and left; she also induced her sister to falsely accept responsibility for a speeding offense, and arranged an inaccurate alibi at trial, leading to A's conviction. The police later searched the sister's empty apartment based on a prosecutor's (not a judge's) order after detecting marijuana odor, found narcotics; the admissibility of that evidence is disputed.
& 
Part I: Determine the criminal liability of A, her sister, and her friend. Part II: Analyze the admissibility of the narcotics evidence and whether an objection is required for any exclusion.\\
\hline
\textbf{Public Law:} A city enacted and published a pigeon-feeding ban that was adopted by the council despite not being on the council's meeting agenda. After a resident kept feeding pigeons, the city issued an individual enforcement order; he timely sued to annul it, alleging defects in the ordinance's adoption and legal basis. 
&
Assess the likelihood of success of the annulment action against the individual order, including incidental review of the underlying ordinance's validity.\\
\hline
\end{tabular}

\label{tab:case_facts_tasks}
\end{table}

\begin{table}[htbp]
\centering
\caption{Summary statistics for criminal and public law. Grades use a 0--18 scale. QWK (public law) between the instructor's grade and the rubric-equipped assistants.  Dist. uses bins: \textbf{[0--3], [4--6], [7--9], [10--12], [13--15], [16--18]}(counts)~\cite{FUberlinLaw_grading_scales}. 
}
\begin{tabular}{@{}lcc@{}}
\toprule
Dataset & \textbf{Criminal Law} & \textbf{Public Law} \\
\midrule
Setting & Take-home & Exam cond. \\
Grade ($\mu^{\pm\sigma}$; min--max) & \nsd{7.437}{3.569}; 2--16 & \nsd{6.438}{3.932}; 1--15 \\
Dist. (bins) & 8/25/22/8/6/2 & 5/3/3/4/1/0 \\
Pass rate  (\%) & 77.46 & 68.75 \\
QWK ($\mu^{\pm\sigma}$) & N/A &\nsd{0.911}{0.066} \\
Tokens/words  ($\mu^{\pm\sigma}$) & \nsd{11796}{1639}/\nsd{6573}{878} & \nsd{2708}{781}/\nsd{1481}{447} \\
Rubric/Soln. & Yes/Yes & Yes/Yes  \\
N & 71 & 16 \\
\bottomrule
\end{tabular}
\label{tab:dataset_stats}
\end{table}

\subsection{Model Selection}

\begin{table}[htbp]
\centering
\caption{LLMs used in our experiments, categorized by reasoning capability. \colorbox{gray!15}{Shaded cells}: open-source; Non-shaded: closed-source.}
\label{tab:models}

\begin{tabular}{p{0.20\linewidth}p{0.71\linewidth}}
\toprule
\textbf{Category} & \textbf{LLMs} \\
\midrule

\textbf{Reasoning} &
GPT-5; GPT-5-mini~\cite{openai_gpt5_system_card_2025};
GPT-5.1~\cite{openai_gpt51_system_card_addendum_2025};
GPT-5.2~\cite{openai_gpt52_system_card_update_2025};
Gemini-2.5-Pro~\cite{google_gemini25_thinking_updates_2025}
\\

 &
\cellcolor{gray!15}DeepSeek-685B-V3.1~\cite{deepseekai2024deepseekv3technicalreport};
\cellcolor{gray!15}GPT-OSS-20B, \cellcolor{gray!15}GPT-OSS-120B~\cite{openai2025gptoss120bgptoss20bmodel};
\cellcolor{gray!15}Qwen3-235B-Th,
\cellcolor{gray!15}Qwen3-30B-Th,
\cellcolor{gray!15}Qwen3-Next-80B-Th,
\cellcolor{gray!15}Qwen3-32B~\cite{qwen3technicalreport,qwen_blog_4074cca8};
\cellcolor{gray!15}QwQ-32B~\cite{qwq32b};
\cellcolor{gray!15}Ministral-3-14B-Rea~\cite{liu2026ministral3}
\\

\midrule

\textbf{Non-reasoning} &
GPT-4o,
GPT-4o-mini~\cite{openai_hello_gpt4o_2024, gpt4o_mini_openai};
GPT-4.1,
GPT-4.1-mini~\cite{openai_gpt41_introduction_2025}
\\

 &
\cellcolor{gray!15}Mistral-Large-3-675B~\cite{mistral_3_announcement};
\cellcolor{gray!15}Gemma-3-27B-it~\cite{gemmateam2025gemma3technicalreport};
\cellcolor{gray!15}Llama-3.3-70B-it~\cite{llama3_3_model_card};

\cellcolor{gray!15}Qwen3-235B,
\cellcolor{gray!15}Qwen3-30B-it,
\cellcolor{gray!15}Qwen3-Next-80B-it~\cite{qwen3technicalreport};
\cellcolor{gray!15}Ministral-3-14B-it~\cite{liu2026ministral3};
\cellcolor{gray!15}Apertus-70B-it~\cite{apertus_70b_instruct_2509};
\cellcolor{gray!15}EuroLLM-22B-it~\cite{eurollm_22b_instruct_2512}
\\

\bottomrule
\end{tabular}
\end{table}

We hypothesize that grading long-form legal case solutions may benefit LLMs from explicit reasoning capabilities. Therefore, we contrast \textit{reasoning} and \textit{non-reasoning} LLMs in our experiments. We also include both open- and closed-source models; open-source models provide greater transparency and stronger privacy guarantees than closed-source alternatives~\cite{10.1145/3708821.3733888}. Our model selection is based on prior work on legal reasoning and grading, prioritizing strong performers while maintaining diversity~\cite{fan2025lexambenchmarkinglegalreasoning,11272818,chlapanis-etal-2025-greekbarbench}. Table~\ref{tab:models} summarizes the evaluated models.

\subsection{Experimental Setup}

To assess how incrementally adding task-specific information in prompts influences grading performance across LLMs, we evaluate four prompting strategies: (1) a task-agnostic (\textit{Task-Agnostic}) prompt with minimal instruction to grade student answers; (2) task-specific instruction plus the grading rubric (\textit{Instr.+Rubric}); (3) task-specific instruction plus a sample solution (\textit{Instr.+Solution}); and (4) task-specific instruction, a sample solution, and a grading rubric (\textit{Instr.+Rubric+Solution}). To ensure robustness across runs and domains, we ran each LLM three times per student-answer for each prompt in both the criminal and public law datasets.
\subsubsection{Prompting Strategies}
\paragraph{Task-Agnostic} As we expect minimal guidance to be insufficient for this task’s complexity, we utilize a task-agnostic prompt as a baseline and expect it to perform worst across LLMs. This setting tests whether LLMs can grade using only their internal knowledge of law exam grading conventions. If models succeed under these constraints, they would support out-of-the-box use without additional prompt engineering, benefiting instructors who want to deploy them as-is. The task-agnostic prompt provides only minimal generic guidance for grading student answers and restricts the model to the grading range and the required output format.

\paragraph{Instr.+Rubric/Solution} To isolate how different types of task information affect grading, we compare prompts that provide either a grading rubric or a sample solution alongside task-specific instruction. This design allows us to assess which information improves grading more when the other is held constant. Thus, for the second prompt (\textit{Instr.+Rubric}), we provide task-specific instruction and the grading rubric but no sample solution. The instructions follow the course instructor's guidance for human graders and are substantially equivalent to the original correction guidelines in terms of legal content and evaluation methodology. Conversely, for \textit{Instr.+Solution}, we replace the rubric with one instructor-written sample solution for the specific case to be solved, keeping the task-specific instruction but providing no rubric.

\paragraph{Instr.+Rubric+Solution} We provide all available grading information in this setting: task-specific instruction, a grading rubric, and the sample solution. Since this condition provides the most complete grading context, we hypothesize that LLMs will perform most effectively in this setting.

Across both datasets, prompts share the same structure and differ only in whether they include task-specific instructions, the rubric, and/or the sample solution. Due to space constraints, we provide the full public-law prompt in the online supplementary\footnote{\url{https://github.com/abdullahalzubaer/icail2026/tree/master/prompts/public_law}}. Criminal-law prompt follows the same structure but cannot be released for data-protection and copyright constraints.

\subsubsection{Evaluation Metric}
\label{subusbsection:evaluation_metric}

Given that grading involves ordinal scores and we want disagreements to be penalized more when grades are farther apart, we evaluate LLM grading performance using the Quadratic Weighted Kappa (QWK), which quantifies inter-rater agreement for ordinal scores. QWK ranges from -1 to 1, where 1 indicates perfect agreement, and -1 indicates agreement worse than chance. It applies quadratic weights, penalizing larger score gaps more than near misses, which is ideal for grading. QWK is also widely used in automated grading~\cite{11272818, li-ng-2024-automated}. 
Following~\cite{11272818}, we compute QWK between each of the three iterations and the expert grade. We then apply a Fisher transformation to the three QWK values, average them in the transformed space, and back-transform to obtain a single Fisher-averaged mean QWK. We report this mean and its standard deviation.
As LLM grade predictions may vary by small numeric amounts across runs while still preserving a strong correlation with expert grade, we also examine Pearson correlation as a complementary metric. However, QWK remains our primary evaluation metric.

\subsubsection{Ensemble Grading}

Beyond evaluating graders as standalone models, we ask: \textit{can combining multiple strong graders yield a grader that is better than the best of its members?}
To answer this question, we compare each ensemble, under a fixed aggregation rule, against the best single member, defined as the highest QWK achieved by any member when evaluated standalone on the same prompt variant.

We form two open-source ensembles.
\textit{Open-L}: highest performing large open models. \textit{Open-S:} moderate-performing smaller open models restricting model size to 20B-80B parameters).
Each ensemble includes three models, selected by average performance across criminal and public law.
We focus on open-source models because they are cheaper to run and easier to deploy in settings where cost and privacy matter.

For each student answer and iteration $i \in {1,2,3}$, we aggregate the three predicted grades using two fixed operators: (1) following prior work that uses conservative aggregation~\cite{fan2025lexambenchmarkinglegalreasoning}, and (2) Median, a robust estimator that is less sensitive to outliers and calibration differences across members.
We report the minimum because it is commonly used, but we do not assume it is theoretically well-founded in our setting: if members exhibit different score ranges or systematic biases, the minimum can be dominated by the harshest grader and introduce additional bias.
Median aggregation directly targets this failure mode by reducing sensitivity to a single pessimistic (or overly optimistic) member.
We then compute QWK for each ensemble iteration relative to the expert grade as described in Section~\ref{subusbsection:evaluation_metric}.

To make the ensemble hypothesis explicit, we report each ensemble score alongside ``Best-Member'' under the same prompt variant (Table~\ref{tab:ensemble_vs_bestconstituent_marker}) and interpret improvements as evidence that aggregation can surpass the strongest member model.

\subsubsection{Implementation Details} 

We primarily accessed LLMs via OpenRouter~\cite{openrouter_homepage}, using LiteLLM's Chat Completion endpoint~\cite{litellm_completion_docs} to standardize API calls within a single interface. OpenAI's closed models were accessed directly via OpenAI's Chat Completions endpoint~\cite{openai_chat_api_docs}.
For models that were either unavailable via OpenRouter or more computationally efficient to run locally\footnote{Qwen3-30B-it, Qwen3-30B-Th., Qwen3-32B, QwQ-32B, Ministral-3-14B-it, Ministral-3-14B-Rea., Gemma-3-27b-it, Apertus-70b-it, EuroLLM-22b-it}, we performed inference with vLLM~\cite{10.1145/3600006.3613165} on up to four NVIDIA A100 GPUs, depending on model size. We used the default hyperparameters recommended by each provider for all LLMs, consistent with prior work that adopts default configuration (e.g.,~\cite{chlapanis-etal-2025-greekbarbench}). These configurations are suggested by the model developers and are intended to yield optimal performance. 
Following common benchmarking practice (e.g.,~\cite{chlapanis-etal-2025-greekbarbench}) and due to computational and monetary constraints (especially for closed-source models), we ran three full-dataset iterations per LLM to obtain a coarse estimate of run-to-run variability.
To handle failures during generation or outputs that did not conform to the expected format or constraints, we allowed up to 20 retries for model outputs and up to 10 retries for temporary execution failures. In practice, fewer than 10 retries were sufficient in most cases. Using OpenRouter pricing, the total inference cost for successful generations was $\approx$\$280 (criminal law: \$247; public law: \$33) (excluding OpenAI-direct usage, which is only available as account-level aggregates, and excluding unlogged failed retries).

\section{Results}

\begin{table*}[htbp]
  \centering

\caption{QWK scores for prompt variants reported as Criminal $\mid$ Public law. Within each block-Reasoning / Non-Reasoning;
\textbf{bold} indicates the best and \underline{underline} the runner-up, computed separately for criminal and public per prompt variants.
\colorbox{gray!15}{Shaded cells}: open-source LLMs; Non-shaded: closed-source LLMs.
Values are reported as \nsd{\textbf{QWK}}{\textbf{SD}} over three runs. Prompt-invariant baselines (incl. prior work where available) appear as table notes$^{\mathrm{a,b}}$. \dag Best model.}

  \label{results:tab:all_prompts} 

  \begin{tabular}{l|c|c|c|c@{}}
    \toprule
    \textbf{LLMs} &
    \makecell{\textbf{Task Agnostic}\\\textbf{\underline{Criminal} $\mid$ Public}} & 
    \makecell{\textbf{Instr.+Rubric}\\\textbf{Criminal $\mid$ Public}} & 
    \makecell{\textbf{Instr.+Solution}\\\textbf{Criminal $\mid$ Public}} & 
    \makecell{\textbf{Instr.+Rubric+Sol}\\\textbf{Criminal $\mid$ Public}} \\ 
    
    \midrule
    \multicolumn{5}{c}{\textit{Reasoning}} \\
    \midrule

    GPT-5.2  &
    $\qwk{\underline{0.311}}{0.049} \mid \qwk{0.356}{0.056}$ & 
    $\qwk{0.298}{0.036} \mid \qwk{0.685}{0.068}$ & 
    $\qwk{{\textbf{0.465}}}{0.036} \mid \qwk{0.621}{0.041}$  & 
    $\qwk{0.431}{0.016} \mid \qwk{0.873}{0.049}$  \\ 

    GPT-5$_{\text{}}^{\textbf{\dag}}$  &
    $\qwk{\textbf{0.309}}{0.017} \mid \qwk{\underline{0.576}}{0.030}$ & 
    $\qwk[]{\textbf{0.557}}{0.050} \mid \qwk{\underline{0.846}}{0.045}$ & 
    $\qwk{{0.392}}{0.019} \mid \qwk{\textbf{0.793}}{0.038}$  & 
    $\qwk[]{\textbf{0.599}}{0.005} \mid \qwk[]{\textbf{0.911}}{0.016}$  \\ 

    GPT-5.1$_{\text{}}$ &
    $\qwk{0.181}{0.016} \mid \qwk{\textbf{0.596}}{0.043}$ & 
    $\qwk{0.145}{0.004} \mid \qwk{0.399}{0.064}$ & 
    $\qwk{0.151}{0.019} \mid \qwk{0.470}{0.037}$  & 
    $\qwk{0.086}{0.011} \mid \qwk{0.361}{0.044}$  \\ 

    Gemini-2.5-Pro$_{\text{}}$ &
    $\qwk{0.161}{0.030} \mid \qwk{0.568}{0.149}$ & 
    $\qwk{0.475}{0.011} \mid \qwk{\textbf{0.860}}{0.048}$ & 
    $\qwk{{\underline{0.430}}}{0.017} \mid \qwk{\underline{0.676}}{0.051}$  & 
    $\qwk{\underline{0.565}}{0.015} \mid \qwk[]{\underline{0.884}}{0.008}$  \\ 

    GPT-5-mini$_{\text{}}$ &
    $\qwk{0.097}{0.024} \mid \qwk{0.224}{0.098}$ & 
    $\qwk{\underline{0.499}}{0.047} \mid \qwk{0.697}{0.038}$ & 
    $\qwk{0.270}{0.033} \mid \qwk{0.558}{0.022}$  & 
    $\qwk{0.425}{0.004} \mid \qwk{0.641}{0.046}$  \\ 

    \cellcolor{gray!15}DeepSeek-685B-V3.1$_{\text{}}$ &
    $\qwk{0.088}{0.046} \mid \qwk{0.501}{0.181}$ & 
    $\qwk{0.287}{0.089} \mid \qwk{0.581}{0.190}$ & 
    $\qwk{0.283}{0.093} \mid \qwk{0.600}{0.052}$  & 
    $\qwk{0.277}{0.05} \mid \qwk{0.785}{0.026}$  \\ 

    \cellcolor{gray!15}GPT-OSS-20B$_{\text{}}$&
    $\qwk{0.065}{0.098} \mid \qwk{0.073}{0.238}$ & 
    $\qwk{0.250}{0.084} \mid \qwk{0.628}{0.132}$ & 
    $\qwk{0.171}{0.036} \mid \qwk{0.538}{0.090}$  & 
    $\qwk{0.282}{0.112} \mid \qwk{0.712}{0.084}$  \\ 

    \cellcolor{gray!15}Qwen3-Next-80B-Th.$_{\text{}}$ &
    $\qwk{0.050}{0.011} \mid -\qwk{0.049}{0.129}$ & 
    $\qwk{0.309}{0.006} \mid \qwk{0.635}{0.125}$ & 
    $\qwk{0.269}{0.060} \mid \qwk{0.527}{0.044}$  & 
    $\qwk{0.336}{0.016} \mid \qwk{0.789}{0.052}$  \\ 

    \cellcolor{gray!15}Ministral-3-14B-Rea.$_{\text{}}$ &
    $\qwk{0.030}{0.010} \mid \qwk{0.036}{0.022}$ & 
    $\qwk{0.072}{0.040} \mid -\qwk{0.007}{0.294}$ & 
    $\qwk{0.052}{0.036} \mid \qwk{0.095}{0.049}$  & 
    $-\qwk{0.008}{0.109} \mid \qwk{0.287}{0.048}$  \\ 

    \cellcolor{gray!15}Qwen3-235B-Th.$_{\text{}}$ &
    $\qwk{0.008}{0.011} \mid \qwk{0.153}{0.042}$ & 
    $\qwk{0.305}{0.032} \mid \qwk{0.604}{0.016}$ & 
    $\qwk{0.234}{0.005} \mid \qwk{0.450}{0.136}$  & 
    $\qwk{0.453}{0.034} \mid \qwk{0.787}{0.046}$  \\ 

    \cellcolor{gray!15}Qwen3-32B$_{\text{}}$ &
    $\qwk{0.000}{0.007} \mid \qwk{0.006}{0.030}$ & 
    $\qwk{0.042}{0.033} \mid \qwk{0.156}{0.201}$ & 
    $\qwk{0.065}{0.038} \mid \qwk{0.122}{0.035}$  & 
    $\qwk{0.117}{0.053} \mid \qwk{0.136}{0.134}$  \\ 

    \cellcolor{gray!15}Qwen3-30B-Th. &
    $\qwk{-0.007}{0.016} \mid \qwk{-0.015}{0.036}$ & 
    $\qwk{0.129}{0.011} \mid \qwk{0.505}{0.045}$ & 
    $\qwk{0.101}{0.023} \mid \qwk{0.461}{0.061}$  & 
    $\qwk{0.306}{0.041} \mid \qwk{0.506}{0.094}$  \\ 

    \cellcolor{gray!15}QwQ-32B &
    $\qwk{-0.008}{0.028} \mid \qwk{0.080}{0.083}$ & 
    $\qwk{0.068}{0.033} \mid \qwk{0.146}{0.085}$ & 
    $\qwk{0.083}{0.012} \mid \qwk{0.367}{0.053}$  & 
    $\qwk{0.050}{0.023} \mid \qwk{0.343}{0.061}$  \\ 

    \cellcolor{gray!15}GPT-OSS-120B&
    $\qwk{-0.032}{0.054} \mid \qwk{-0.085}{0.103}$ & 
    $\qwk{0.284}{0.02} \mid \qwk{0.400}{0.100}$ & 
    $\qwk{0.307}{0.027} \mid \qwk{0.469}{0.073}$  & 
    $\qwk{0.353}{0.021} \mid \qwk{0.744}{0.084}$  \\ 

    \midrule
    \multicolumn{5}{c}{\textit{Non-Reasoning}} \\
    \midrule

    \cellcolor{gray!15}Mistral-Large-3-675B &
    $\qwk{\textbf{0.067}}{0.004} \mid \qwk{0.113}{0.017}$ & 
    $\qwk{\underline{0.270}}{0.023} \mid \qwk[]{\textbf{0.709}}{0.059}$ & 
    $\qwk{\underline{0.205}}{0.007} \mid \qwk{\underline{0.423}}{0.016}$  & 
    $\qwk{\underline{0.347}}{0.020} \mid \qwk[]{\textbf{0.790}}{0.045}$  \\ 

    GPT-4o &
    $\qwk{\underline{0.063}}{0.005} \mid \qwk{\underline{0.283}}{0.036}$ & 
    $\qwk{0.114}{0.018} \mid \qwk{0.279}{0.036}$ & 
    $\qwk{0.168}{0.026} \mid \qwk{0.409}{0.057}$  & 
    $\qwk{0.145}{0.022} \mid \qwk{0.466}{0.123}$  \\ 

    GPT-4.1 &
    $\qwk{0.052}{0.008} \mid \qwk{\textbf{0.338}}{0.038}$ & 
    $\qwk[]{\textbf{0.405}}{0.025}\mid \qwk{\underline{0.589}}{0.041}$ & 
    $\qwk{\textbf{0.236}}{0.007} \mid \qwk{\textbf{0.678}}{0.055}$  & 
    $\qwk[]{\textbf{0.478}}{0.025} \mid \qwk{\underline{0.659}}{0.074}$  \\ 

    \cellcolor{gray!15}Ministral-3-14B-it &
    $\qwk{0.022}{0.001} \mid \qwk{0.013}{0.008}$ & 
    $\qwk{0.058}{0.037} \mid \qwk{0.538}{0.098}$ & 
    $\qwk{0.064}{0.007} \mid \qwk{0.375}{0.068}$  & 
    $\qwk{0.113}{0.025} \mid \qwk{0.514}{0.155}$  \\ 

    \cellcolor{gray!15}Gemma-3-27B-it &
    $\qwk{0.015}{0.007} \mid \qwk{0.011}{0.004}$ & 
    $\qwk{0.124}{0.032} \mid \qwk{0.053}{0.004}$ & 
    $\qwk{0.011}{0.007} \mid \qwk{0.060}{0.021}$  & 
    $\qwk{0.075}{0.081} \mid \qwk{0.044}{0.028}$  \\ 

    GPT-4.1-mini &
    $\qwk{0.013}{0.001} \mid \qwk{0.096}{0.029}$ & 
    $\qwk{0.156}{0.004} \mid \qwk{0.321}{0.084}$ & 
    $\qwk{0.103}{0.018} \mid \qwk{0.176}{0.049}$  & 
    $\qwk{0.149}{0.043} \mid \qwk{0.598}{0.084}$  \\ 

    GPT-4o-mini &
    $\qwk{0.013}{0.004} \mid \qwk{0.037}{0.013}$ & 
    $\qwk{-0.075}{0.064} \mid \qwk{0.017}{0.058}$ & 
    $\qwk{0.087}{0.029} \mid \qwk{0.034}{0.216}$  & 
    $\qwk{0.107}{0.094} \mid \qwk{0.113}{0.013}$  \\ 

    \cellcolor{gray!15}Qwen3-Next-80B-it &
    $\qwk{0.010}{0.004} \mid \qwk{0.074}{0.013}$ & 
    $\qwk{0.030}{0.013} \mid \qwk{0.262}{0.032}$ & 
    $\qwk{0.056}{0.008} \mid \qwk{0.200}{0.036}$  & 
    $\qwk{0.052}{0.03} \mid \qwk{0.263}{0.038}$  \\ 

    \cellcolor{gray!15}Qwen3-235B &
    $\qwk{0.007}{0.005} \mid \qwk{0.026}{0.007}$ & 
    $\qwk{0.012}{0.061} \mid \qwk{0.456}{0.176}$ & 
    $\qwk{0.054}{0.018} \mid \qwk{0.206}{0.07}$  & 
    $\qwk{0.096}{0.017} \mid \qwk{0.366}{0.086}$  \\ 

    \cellcolor{gray!15}Qwen3-30B-it. &
    $\qwk{0.006}{0.006} \mid \qwk{0.001}{0.016}$ & 
    $\qwk{0.052}{0.049} \mid \qwk{0.052}{0.284}$ & 
    $\qwk{0.036}{0.008} \mid \qwk{0.092}{0.029}$  & 
    $\qwk{0.082}{0.064} \mid \qwk{0.104}{0.082}$  \\ 
    
    \cellcolor{gray!15}Apertus-70B-it &
    $\qwk{0.005}{0.005} \mid \qwk{-0.004}{0.047}$ & 
    $\qwk{0.117}{0.206} \mid \qwk{-0.002}{0.241}$ & 
    $\qwk{0.058}{0.057} \mid \qwk{0.037}{0.209}$  & 
    $\qwk{0.002}{0.074} \mid \qwk{0.314}{0.292}$  \\ 

    \cellcolor{gray!15}Llama-3.3-70B-it &
    $\qwk{0.004}{0.004} \mid \qwk{0.005}{0.017}$ & 
    $\qwk{0.074}{0.035} \mid \qwk{0.175}{0.137}$ & 
    $\qwk{0.027}{0.017} \mid \qwk{0.118}{0.017}$  & 
    $\qwk{0.062}{0.039} \mid \qwk{0.172}{0.102}$  \\ 

    \cellcolor{gray!15}EuroLLM-22B-it &
    $\qwk{-0.015}{0.015} \mid \qwk{0.019}{0.041}$ & 
    $\qwk{-0.007}{0.072} \mid \qwk{-0.050}{0.056}$ & 
    $\qwk{0.037}{0.006} \mid \qwk{0.066}{0.049}$  & 
    $\qwk{-0.124}{0.104} \mid \qwk{-0.031}{0.285}$  \\ 

\midrule
\multicolumn{5}{@{}l@{}}{%
\footnotesize\parbox{0.98\linewidth}{%
$^{\mathrm{a}}$ \textbf{Public.}
(1) \textit{Human:} highest agreement between human graders, $\qwk{0.911}{0.066}$.
(2) \textit{Gemini 2.5 Pro (prior work):} QWK $0.764$, from~\cite{strecker2026ki}.\\
$^{\mathrm{b}}$ \textbf{Random baseline.}
(Criminal $\mid$ Public): $\qwk{-0.053}{0.260} \mid \qwk{0.068}{0.365}$.%
}} \tabularnewline
  \end{tabular}
\end{table*}

In Table~\ref{results:tab:all_prompts} we report QWK between LLM-assigned grades and expert reference grades on the 18-point German scale. We evaluated prompting strategies that progressively add task information: Task-Agnostic, Instr.+Rubric, Inst.+Solution, and Instr.+Rubric+Solution. Table~\ref{results:tab:all_prompts} is organized into two blocks: reasoning and non-reasoning. Overall, performance improves monotonically with additional task information. Strongest result obtained when the model has access to both the reference solution and the grading rubric.

\subsection{Overall grading performance and model comparison}
Under the most informative prompt (Instr.+Rubric+Solution), reasoning LLMs achieve the highest agreement with expert graders. In public law, the best single-model result is 0.911 for GPT-5, matching human agreement (0.911; see Table~\ref{results:tab:all_prompts}, note) when the rubric and solution are provided, which also exceeds prior work (0.764)~\cite{strecker2026ki}. In criminal law, performance differs: under full context, the best result is 0.599 for GPT-5, substantially below the public law score for the same model. This gap persists across model families and prompting conditions. In criminal law, however, nearly all models in both the reasoning and non-reasoning blocks exceed previously reported results~\cite{wendlinger2024suitabilitypretrainedfoundationalllms} (Table~\ref{appendix:tab:criminal_law_correlation_lorenz} in Appendix~\ref{appendix}), with the largest gains under prompts that add task information.
The runner-up models also perform worse in criminal law than in public law (e.g., Gemini-2.5-Pro: 0.565 vs. 0.884).
Why might this be the case? We hypothesize that part of the gap reflects differences in task complexity induced by the underlying exam formats (Section~\ref{dataset_overview}). In particular, the criminal law dataset originates from a more open-ended setting, which tends to yield longer and more varied solution structures than time-bound public law mock exams. These differences likely contribute to lower LLM agreement in criminal law under otherwise comparable scoring setups. As a robustness check, we also examined Pearson correlation. The Pearson-based results are broadly consistent with the QWK analysis. In criminal law, the top-5 model set is identical under both metrics, differing only in order. In public law, four of the top five models overlap, and GPT-5 ranks first under both and remains the best model. The strongest results under both metrics are also predominantly obtained with the most informative prompt setting.

Across \textit{model classes}, top performance (best and runner-up) is achieved by reasoning models rather than non-reasoning models.
The best performance in non-reasoning models is behind the runner-ups of reasoning models for each prompt variant in both datasets (except GPT-4.1).
For example, under full context, strong reasoning models such as GPT-5 (0.911/0.599 in criminal/public law) and Gemini-2.5-Pro (0.884/0.565) outperform non-reasoning baselines.
The top non-reasoning model under full context reaches 0.790 in public law (Mistral-Large-3-675B) and 0.478 in criminal law (GPT-4.1).
We observe a similar pattern in open-source models as well, for example, in Qwen3-Next-80B and Qwen3-235B, except for the Ministral3 series.
This also indicates that for the more complex criminal law solution, reasoning is required, while public law grading is possible with less verbosity, and therefore, more efficient models.

Turning to \textit{open source} LLMs, Mistral-Large-3-675B performs best overall.
It outperforms all reasoning and non-reasoning models in both datasets, except Qwen3-235B-Th and GPT-OSS-120B, in criminal law. 
However, reasoning model Qwen3-Next-80B-Th. reaches 0.789/0.336 in criminal/public law. This is slightly below Mistral-Large-3-675B (0.790/0.347)
while using about 13$\times$ fewer active parameters (3B vs 41B).
This suggests that reasoning models can offer a favorable performance-efficiency trade-off while being computationally efficient.
Overall, open-source models can be competitive, but they generally lag behind the closed-source ceiling.

Not all open-source or domain-adapted models are viable graders in our setting. Among fully open models, EuroLLM~\cite{eurollm_22b_instruct_2512} and Apertus~\cite{apertus_70b_instruct_2509} achieved very low agreement across prompt settings. We also evaluated domain-specific legal models. Saul-54B~\cite{NEURIPS2024_ea3f85a3} could not be applied to the criminal law scenario due to context length limitations. In public law, it remained low-performing (best score: 0.158 under the Instr.+Rubric). Saul-7B-Instruct~\cite{colombo2024saullm7bpioneeringlargelanguage} and leo-hessianai-70b-chat~\cite{leolm_70b_chat_hf} frequently failed to follow the required output format.

\subsection{Prompt ablation: what information is necessary?}

Table~\ref{results:tab:all_prompts} shows that task-agnostic prompting is not sufficient for reliable grading of German legal exams.
Under a task-agnostic prompt, several models achieve near-zero or even negative QWK, particularly in public law; smaller open-source models are close to chance-level agreement (see note in Table~\ref{results:tab:all_prompts}).
This indicates that ``out of the box'' grading without task-specific information is not useful for grading law exams. 

Adding task-specific information substantially improves agreement{(Figure~\ref{appendix:performance_gain_prompt_ablation} in Appendix~\ref{appendix})}. Both the rubric and the sample solution are beneficial; however, on average, the grading rubric plays a more prominent role for all the models.
For instance, reasoning closed-source model GPT-5 improves from 0.576 to  0.846 in public law from Task-Agnostic to Instr.+rubric prompt. Whereas, Instr.+Solution yields a smaller gain (0.793). A similar pattern is observed in criminal law as well for the same model (0.309 to 0.557 with rubric vs 0.392 with solution).
This trend also holds for the non-reasoning open-source model Mistral-Large-3-675B.
In public law, performance increases from 0.113 to 0.709 with the rubric versus 0.423 with the solution, and in criminal law from 0.067 to 0.270 with the rubric versus 0.205 with the solution.
This suggests that an explicit mapping from legal quality to grade via a rubric provides critical information for grading decisions, beyond having access to an idealized reference answer alone.
We view this as a positive result because rubrics are lower effort to design than providing complete sample solutions.

Finally, providing both the grading rubric and the solution produces the most consistent peak performance across models and domains. Under Instr.+Rubric+Solution, top reasoning models demonstrated high agreement with relatively low within-run variance. For example, GPT-5 with \nsd{0.911}{0.016} in public law, and \nsd{0.599}{0.005} in criminal law. A similar pattern is also present in other models, for example, Qwen3-Next-80B-Th, GPT-OSS-20/120B, and Mistral-Large-3-675B.
These results indicate that LLM grading is most effective when the model has access to both the grading rubric and the sample solution.

\subsection{Ensemble grading}

\begin{table*}[t]
  \centering

\caption{Ensemble grading performance (\nsd{\textbf{QWK}}
{\textbf{SD}}) under the same prompt variant, using min and median aggregation. \textbf{Best-member} is the highest-QWK single ensemble member evaluated standalone. \textbf{Bold} denotes ensembles exceeding Best-Member on the same dataset, and \underline{\textbf{bold}} marks the best ensemble per dataset within each ensemble group over all prompt variants.}
  \label{tab:ensemble_vs_bestconstituent_marker}

  \begin{tabular}{@{}l|l|c|c|c|c@{}}
    \toprule
    \textbf{Ensemble} &
    \makecell{\textbf{Agg.}\\\textbf{/\textit{Member}}} &
    \makecell{\textbf{Task-Agnostic}\\\textbf{Criminal $\mid$ Public}} &
    \makecell{\textbf{Instr.+Rubric}\\\textbf{Criminal $\mid$ Public}} &
    \makecell{\textbf{Instr.+Solution}\\\textbf{Criminal $\mid$ Public}} &
    \makecell{\textbf{Instr.+Rubric+Solution}\\\textbf{Criminal $\mid$ Public}} \\
    \midrule

    \multirow{3}{*}{{\textit{Open-L}}} &
    {Min} &
    $\qwk{\textbf{0.104}}{0.037} \mid \qwk{0.499}{0.180}$ &
    $\qwk{\textbf{0.359}}{0.074} \mid \qwk{0.657}{0.153}$ &
    $\qwk{\textbf{0.319}}{0.094} \mid \qwk{0.599}{0.057}$ &
    $\qwk{{0.388}}{0.030} \mid \qwk{0.773}{0.020}$ \\
    &
    {Median } &
    $\qwk{0.062}{0.020} \mid \qwk{0.144}{0.014}$ &
    $\qwk{\textbf{0.324}}{0.025} \mid \qwk{\textbf{0.740}}{0.068}$ &
    $\qwk{0.267}{0.033} \mid \qwk{0.494}{0.068}$ &
    $\qwk{\textbf{\underline{0.467}}}{0.018} \mid \qwk{\textbf{\underline{0.881}}}{0.027}$ \\
    &
    {\textit{Best-Member}} &
    $\qwk{0.088}{0.046} \mid \qwk{{0.501}}{0.181}$ &
    $\qwk{0.305}{0.032} \mid \qwk{0.709}{0.059}$ &
    $\qwk{0.283}{0.093} \mid \qwk{{0.600}}{0.052}$ &
    $\qwk{0.453}{0.034} \mid \qwk{0.790}{0.045}$ \\
    \midrule

    \multirow{3}{*}{{\textit{Open-S}}} &
    {Min} &
    $\qwk{0.029}{0.076} \mid \qwk{-0.013}{0.206}$ &
    $\qwk{\textbf{0.322}}{0.072} \mid \qwk{0.617}{0.091}$ &
    $\qwk{0.214}{0.043} \mid \qwk{\textbf{0.673}}{0.090}$ &
    $\qwk{\textbf{\underline{0.488}}}{0.077} \mid \qwk{\textbf{\underline{0.829}}}{0.108}$ \\
    &
    {Median} &
    $\qwk{0.043}{0.010} \mid \qwk{{0.023}}{0.184}$ &
    $\qwk{0.282}{0.018} \mid \qwk{\textbf{0.671}}{0.042}$ &
    $\qwk{0.257}{0.061} \mid \qwk{\textbf{0.578}}{0.014}$ &
    $\qwk{\textbf{0.365}}{0.040} \mid \qwk{0.767}{0.036}$ \\
    &
    {\textit{Best-Member}} &
    $\qwk{0.065}{0.098} \mid \qwk{0.073}{0.238}$ &
    $\qwk{0.309}{0.006} \mid \qwk{0.635}{0.125}$ &
    $\qwk{0.269}{0.060} \mid \qwk{0.538}{0.090}$ &
    $\qwk{0.336}{0.016} \mid \qwk{0.789}{0.052}$  \\
    \midrule

    \multicolumn{6}{@{}l@{}}{%
      \footnotesize\parbox{0.95\linewidth}{%
      
      \textbf{Complete model names} \textit{Open-L}: Mistral-Large-3-675B, DeepSeek-685B-V3.1, and Qwen3-235B-Th. \textit{Open-S}: GPT-OSS-20B, Qwen3-Next-80B-Th., and Qwen3-30B-Th.%
      }%
    } \tabularnewline

  \end{tabular}
\end{table*}

Table~\ref{tab:ensemble_vs_bestconstituent_marker} compares each ensemble to \textit{Best-Member}, the highest scoring member under the same prompt variant. Across both Open-L and Open-S\footnote{\textit{Open-L}: Mistral-Large-3-675B, DeepSeek-685B-V3.1, and {Qwen3-235B-Th.} \textit{Open-S}: GPT-OSS-20B, Qwen3-Next-80B-Th., and Qwen3-30B-Th.}, the best ensemble exceeds the best single-member score in \emph{both} criminal and public law, showing that ensembling can outperform even its strongest standalone model.

For \textit{Open-L}, median aggregation performs best and yields the largest gains under full context. It exceeds Best-Member under Instr.+Rubric (criminal: 0.359 vs.\ 0.305; public: 0.740 vs.\ 0.709) and improves further under Instr.+Rubric+Solution (0.467 vs.\ 0.453; 0.881 vs.\ 0.790). Although public law does not improve in the task-agnostic setting or under Instr.+Solution.

For \textit{Open-S}, in contrast to Open-L, minimum aggregation performs best, with the strongest results under full context, outperforming Best-Member on criminal and public law (0.488 vs.\ 0.336; 0.829 vs.\ 0.789). No gains appear in the task-agnostic setting, and criminal law does not improve under Instr.+Solution. In addition, open-source ensembling can narrow the gap to our best closed single model (GPT-5) and approach the runner-up (Gemini-2.5-Pro).

Table~\ref{tab:error_bucket_distribution} further compares the error profiles of the strongest proprietary models and the best-performing ensembles under Instr.+Rubric+Solution. In criminal law, the ensembles mainly shift error types: Open-L$_{\mathrm{median}}$ minimizes under-grading ($\leq\!-2$, $-1$), Open-S$_{\mathrm{min}}$ minimizes $+1$ errors, and GPT-5 remains best on large over-grading ($\geq\!+2$). In public law, proprietary models remain stronger overall, with GPT-5 minimizing large under-grading and Gemini~2.5 achieving the highest exact-match rate. This suggests that ensembling mainly rebalances specific error modes rather than producing a uniform advantage across all error categories.

\begin{table}[htbp]
\centering

\caption{Error-bucket distribution for the strongest prompt variant (\textit{Instr.+Rubric+Solution}) for the best and runner-up proprietary models and the two best-performing ensembles. Cells show mean proportion$^{\pm\mathrm{SD}}$. $\uparrow$ higher is better, $\downarrow$ lower is better. \textbf{Bold} marks the best value. }
\label{tab:error_bucket_distribution}
\setlength{\tabcolsep}{2pt}
\small
\begin{tabular}{lccccc}
\toprule
Model & $\leq\!-2\downarrow$ & $-1\downarrow$ & $0\uparrow$ & $+1\downarrow$ & $\geq\!+2\downarrow$ \\
\midrule
\multicolumn{6}{c}{\textit{Criminal Law}} \\
\midrule
GPT-5 & $0.300^{\pm 0.029}$ & $0.127^{\pm 0.037}$ & $0.122^{\pm 0.008}$ & $0.174^{\pm 0.022}$ & $\textbf{0.277}^{\pm 0.016}$ \\
Gemini 2.5 & $0.211^{\pm 0.014}$ & $0.131^{\pm 0.029}$ & $\textbf{0.146}^{\pm 0.016}$ & $0.117^{\pm 0.035}$ & $0.394^{\pm 0.014}$ \\
Open-L$_{\mathrm{med}}$ & $\textbf{0.164}^{\pm 0.016}$ & $\textbf{0.085}^{\pm 0.000}$ & ${0.085}^{\pm 0.014}$ & $0.131^{\pm 0.043}$ & $0.535^{\pm 0.014}$ \\
Open-S$_{\mathrm{min}}$ & $0.296^{\pm 0.086}$ & $0.113^{\pm 0.024}$ & $0.141^{\pm 0.028}$ & $\textbf{0.108}^{\pm 0.035}$ & $0.343^{\pm 0.008}$ \\
\midrule
\multicolumn{6}{c}{\textit{Public Law}} \\
\midrule
GPT-5 & $\textbf{0.062}^{\pm 0.000}$ & $0.042^{\pm 0.036}$ & $0.292^{\pm 0.095}$ & $0.333^{\pm 0.036}$ & $0.271^{\pm 0.095}$ \\
Gemini 2.5 & $0.188^{\pm 0.062}$ & $0.083^{\pm 0.095}$ & $\textbf{0.396}^{\pm 0.180}$ & $0.167^{\pm 0.072}$ & $\textbf{0.167}^{\pm 0.036}$ \\
Open-L$_{\mathrm{med}}$ & $0.208^{\pm 0.036}$ & $\textbf{0.021}^{\pm 0.036}$ & ${0.167}^{\pm 0.072}$ & $0.354^{\pm 0.072}$ & $0.250^{\pm 0.125}$ \\
Open-S$_{\mathrm{min}}$ & $0.375^{\pm 0.108}$ & ${0.104}^{\pm 0.036}$ & $0.208^{\pm 0.072}$ & $\textbf{0.104}^{\pm 0.130}$ & $0.208^{\pm 0.095}$ \\
\bottomrule
\end{tabular}
\end{table}

\section{Discussion and Conclusion}

In this study, we examine whether LLMs can approximate expert grading of long-form German legal case solutions and identify which prompt design enables reliable automated grading.
We evaluate 27 LLMs and find that grading quality depends on task information; task-agnostic prompts perform at near-chance levels.
With full context, reasoning-oriented GPT-5 matches human agreement in public law (0.911) and achieves substantial agreement in criminal law (0.599).
Our approach outperforms previous results~\cite{strecker2026ki,wendlinger2024suitabilitypretrainedfoundationalllms}.
{As our evaluation uses non-public exam answers, these results are less likely to be driven by training-data contamination.}


Turning to the information necessary for grading, we find that providing a structured grading rubric substantially increases alignment with expert grades compared with providing only a sample solution.
This is non-obvious because rubrics often overlap with (and are typically derived from) sample solutions while being less detailed.
Our results, therefore, suggest that \textit{structure, not detail}, drives performance. 
The rubric provides an explicit scoring map from legal quality to the 18-point scale, whereas in the absence of explicit allocation cues, models struggle to translate narrative legal reasoning into the grading scale.
Consistent with this, the rubric advantage persists even when using a simplified criminal law rubric created post hoc by an independent legal expert, despite the sample solution containing richer procedural guidance.
The sample solution is more detailed and gives clearer guidance on what a correct solution should include.
Manual inspection confirms that this concerns both the legal content as well as multiple grader notes on how to deal with mistakes.
It also clarifies which aspects are grading-relevant versus non-relevant ones.
However, models still perform better with the rubric, suggesting that point-referenced structure matters more than an end-to-end exemplar for automated grading in German law exams.
At the same time, agreement peaks when models receive both the rubric and a sample solution, suggesting complementary signals: explicit point allocation from the rubric and exemplar cues for holistic criteria often used in legal exams (e.g., structure and coherence) are most beneficial when combined.
This implies that practitioners should anchor grading to rubric-referenced evidence; adding a sample solution can further improve reliability, but solution-only prompts may be less reliable as score allocation remains implicit.

Beyond prompt content, the choice of model also affects grading reliability.
Not all open-source or domain-adapted legal models are viable graders in our setting, as several show very low agreement or fail due to context length and output-format constraints.
Under identical prompts and full context, reasoning-oriented models achieve higher agreement than non-reasoning models.
This aligns with the need to assess multiple legally coherent reasoning paths, not just surface features.
It also suggests that grading difficulty increases with solution depth and complexity, a pattern that is more pronounced in criminal law.
We also observe a clear domain effect: across the evaluated settings, criminal law yields lower QWK than public law answers.
We interpret this gap as likely linked to exam and data-collection differences (Section~\ref{dataset_overview}), which shape the solution length and depth of legal reasoning required. 
These results indicate that LLM-based grading may serve as a scalable, low-stakes formative tool for practice case feedback and student self-assessment, provided that reliability is established for the target domain and exam format. It should not be used for summative decisions without further validation.


We also find that, beyond single-model grading, ensemble graders offer greater robustness than their member models.
Consistent with the results, the best Open-L and Open-S ensembles outperform the best single member on both criminal and public law.
These gains depend on the aggregation rule and ensemble strength: median aggregation tends to be most reliable for stronger ensembles, consistent with reduced sensitivity to calibration differences.
However, gains are not universal, and some settings show little improvement or even degradation. Error-bucket analysis further suggests that these gains arise less from uniformly reducing all errors than from shifting specific error modes.
This is particularly relevant for privacy-sensitive deployment. Because legal exam answers constitute sensitive educational data, open-source ensembles are a practically important alternative: they can be deployed locally and still narrow the gap to the strongest proprietary models. This makes open-source ensembles the most plausible near-term deployment path in privacy-sensitive educational settings that require local processing of student answers.

\section{Limitations and Future Work}
Taken together, our results highlight some key factors that can drive LLM-based grading performance.
However, our study is limited to German legal exams and a relatively small sample, which limits generalizability across jurisdictions and languages and motivates expanding the dataset. 
In particular, our public law subset is small, and QWK computed on a small dataset can be sensitive to score-distribution effects. Thus, the public law agreement estimate should be treated as preliminary.
More broadly, our results indicate substantial domain dependence, suggesting that transferability to other legal subdomains should be established empirically.
Nevertheless, broader design principles---especially the importance of reasoning-oriented models, rubrics, and sample solutions---may generalize more widely and should be tested in future work across other legal domains.
Our analysis focuses on controlled prompt ablations; future work should test few-shot prompting and stepwise grading pipelines in addition to holistic grading of complete answers.
Our ensembles use only three fixed members and a simple min/median aggregation, which likely limits their improvement over the strongest member.
Extending to alternative ensemble compositions with adaptive aggregation (e.g., calibration-aware weighting) is a promising direction.
We also do not fine-tune models; a larger, curated dataset would enable a systematic study of fine-tuning to improve grading consistency.
Prior work~\cite{11272818} suggests fine-tuned smaller encoder models can outperform general-purpose LLMs on much shorter answers. However, it is unclear whether this extends to long legal solutions and whether the experimental setup would need to be adapted.
Despite these limitations, LLM-based grading may serve as an assistive tool in legal education when deployed with structured task information, careful model selection, and human oversight.

\section*{Code and Data Availability}
\label{data_availability}
Due to data protection laws and copyright constraints, it is not possible to release the student answers. For the criminal law dataset, we also cannot release the prompt or instructor materials. We release our code, the public law prompt, and all instructor materials at \url{https://github.com/abdullahalzubaer/icail2026}.

\section*{GenAI Usage Discloser}
ChatGPT and GitHub Copilot were used during the preparation of this manuscript to assist with code development and with drafting and revising text and tables to improve clarity and readability. All GenAI-assisted outputs were carefully reviewed, verified, edited, and incorporated by the authors as appropriate, who take full responsibility for the final manuscript.

\begin{acks}
The paper has been funded by COMET K1-Competence Center for Integrated Software and AI Systems (INTEGRATE) within the Austrian COMET Program.
\end{acks}

\bibliographystyle{ACM-Reference-Format}
\bibliography{bibliography}

\appendix

\section{Appendix}
\label{appendix}

\begin{figure}[htbp]
  \centering
  \includegraphics[width=0.4\textwidth]{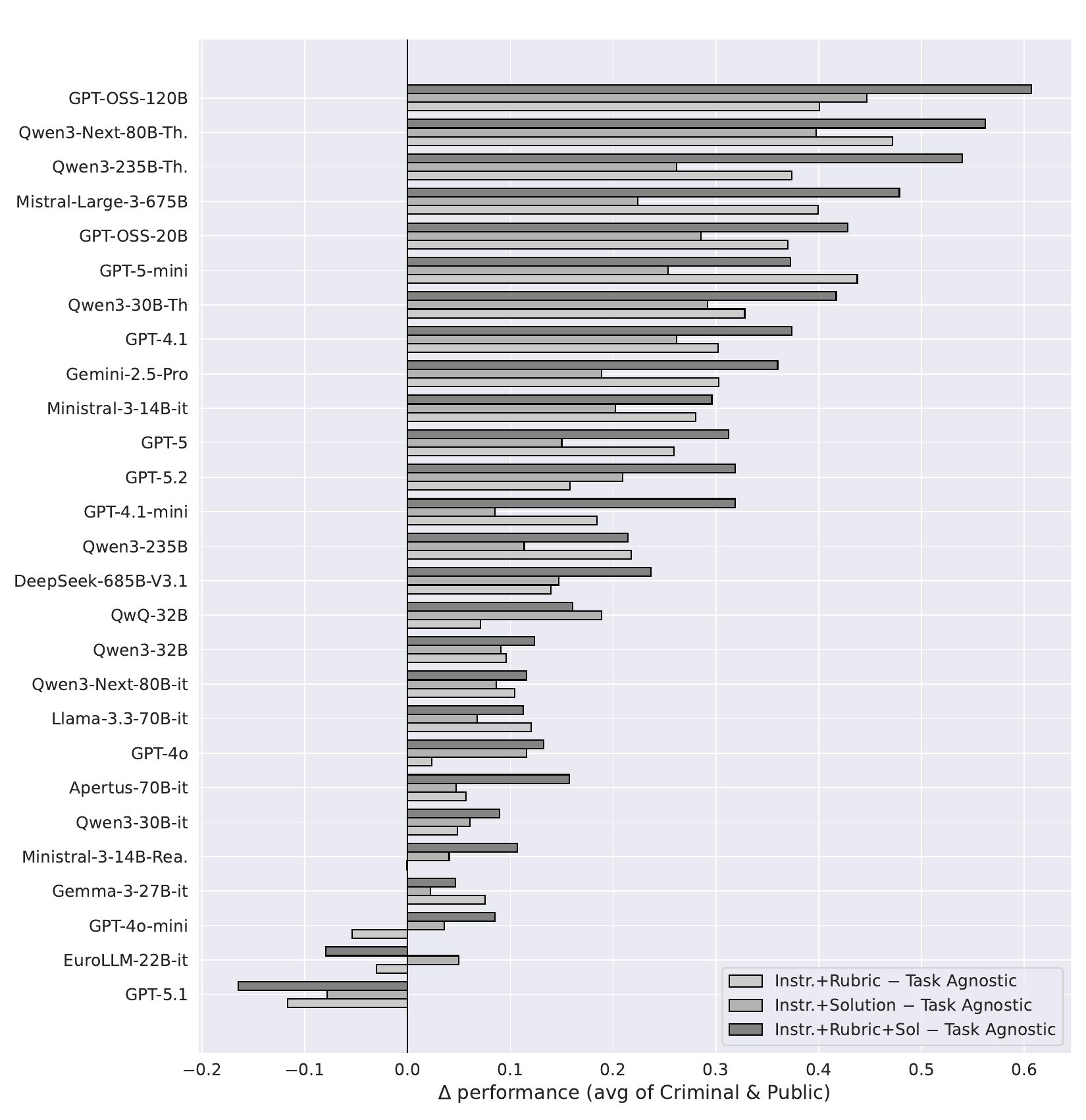}
  \caption{ Performance improvement ($\Delta$) over the Task Agnostic baseline for each model, averaged across the criminal and public dataset, comparing Instr.+Rubric, Instr.+Solution, and Instr.+Rubric+Sol.}

  \label{appendix:performance_gain_prompt_ablation}
\end{figure}

{
\setlength{\tabcolsep}{2pt} 
\begin{table}[htbp]
  \centering
  \footnotesize
\caption{ Agreement across prompt variants for criminal law. \textbf{Bold}/\underline{underline} = best/runner-up per column. Values are \nsd{\textbf{Pearson Corr.}}{\textbf{SD}} over three runs. Top-3 reasoning and non-reasoning models from Table~\ref{results:tab:all_prompts}. \colorbox{gray!15}{Shaded}: open-source LLMs; Non-shaded: closed-source LLMs. Baseline: Corr.$\approx$0.058~\cite{wendlinger2024suitabilitypretrainedfoundationalllms}.}

  \label{appendix:tab:criminal_law_correlation_lorenz}

  \begin{tabular}{@{}p{0.9cm}|l|c|c|c|c@{}}
    \toprule
    &\textbf{LLMs} &
    \makecell{\textbf{Task}\\\textbf{Agnostic}} &
    \makecell{\textbf{Instr.}\\\textbf{+Rubric}} &
    \makecell{\textbf{Instr.}\\\textbf{+Solution}}  &
    \makecell{\textbf{Instr.+Rubric}\\\textbf{+Solution}} \\

    \midrule
\multirow{3}{*}{\rotatebox[origin=c]{90}{\makecell{\small \textbf{Reas}\\ \small \textbf{oning.}}}} &
    GPT-5.1 &
    $\qwk{\textbf{0.497}}{0.045}$ &
    $\qwk{0.446}{0.054}$ &
    $\qwk{0.454}{0.040}$ &
    $\qwk{0.305}{0.026}$ \\

    &GPT-5 &
    $\qwk{\underline{0.454}}{0.031}$ &
    $\qwk{\textbf{0.598}}{0.048}$ &
    $\qwk{0.559}{0.023}$ &
    $\qwk{\textbf{0.599}}{0.005}$ \\

    &GPT-5.2  &
    $\qwk{0.453}{0.069}$ &
    $\qwk{0.481}{0.033}$ &
    $\qwk{\textbf{0.587}}{0.042}$ &
    $\qwk{\underline{0.589}}{0.023}$ \\

\midrule
\multirow{3}{*}{\rotatebox[origin=c]{90}{\makecell{\small \textbf{Non-}\\ \small \textbf{Reas.}}}} &

\cellcolor{gray!15}Mistral-L$^{\mathrm{a}}$ &
    $\qwk{0.429}{0.019}$ &
    $\qwk{0.376}{0.044}$ &
    $\qwk{\underline{0.552}}{0.011}$ &
    $\qwk{0.471}{0.049}$ \\

    &GPT-4o &
    $\qwk{0.228}{0.014}$ &
    $\qwk{0.269}{0.044}$ &
    $\qwk{0.400}{0.031}$ &
    $\qwk{0.325}{0.051}$ \\

    &GPT-4.1  &
    $\qwk{0.369}{0.050}$ &
    $\qwk{\underline{0.581}}{0.030}$ &
    $\qwk{0.492}{0.008}$ &
    $\qwk{0.541}{0.016}$ \\

\midrule
\multicolumn{2}{@{}l@{}}{%
\footnotesize\parbox{0.3\linewidth}{%
$^{\mathrm{a}}$ Mistral-Large-3-675B%
}} \tabularnewline
  \end{tabular}
\end{table}
}

\end{document}